\documentclass[letterpaper,10pt,conference]{ieeeconf}
\IEEEoverridecommandlockouts       
\usepackage{amssymb}
\usepackage{amsmath}
\usepackage[english]{babel}
\usepackage{float}
\usepackage{graphicx}
\usepackage[hidelinks]{hyperref}
\usepackage{mathtools}

\usepackage{filecontents}
\usepackage[noadjust]{cite}
\usepackage[font=footnotesize,labelfont=footnotesize]{subcaption}
\usepackage[font=footnotesize,labelfont=footnotesize]{caption}
\usepackage{comment}
\usepackage{authblk}
\usepackage{multirow}
\usepackage{xcolor}
\usepackage{algorithm}
\usepackage{algorithmic}
\usepackage{xspace}
\usepackage{hhline}

\definecolor{darkblue}{rgb}{0.15,0.15,0.55}
\definecolor{lightgrey}{rgb}{0.75,0.75,0.75}

\providecommand{\codecomment}[1]{\textcolor{lightgrey}{\dotfill}\textcolor{darkblue}{//\,\textrm{#1}}}

\newcommand{\nphard}{$\mathcal{NP}$-hard\xspace}

\newcommand{\thm}{\noindent \textbf{Theorem}\xspace}
\newcommand{\lem}{\noindent \textbf{Lemma}\xspace}
\newcommand{\propo}{\noindent \textbf{Proposition}\xspace}

\newcommand{\pf}{\noindent \textbf{Proof}\xspace}

\newcommand{\qed}{\hfill $\square$}

\begin{document}
\title{\LARGE \bf Coordination of multiple mobile manipulators for ordered sorting of cluttered objects}
\author{Jeeho Ahn$^{1,2}$, Seabin Lee$^1$, and Changjoo Nam$^{1,*}$
\thanks{
This work was supported by Samsung Electronics Co., Ltd (No. IO220810-01912-01). $^1$Dept. of Electronic Eng., Sogang Univ., $^2$Robotics Dept.,  Univ. of Michigan. The present research was conducted while the 1st author was with Sogang. $^*$Corresponding: {\tt\small cjnam@sogang.ac.kr}}
}

\maketitle

\begin{abstract}
We present a coordination method for multiple mobile manipulators to sort objects in clutter. We consider the object rearrangement problem in which the objects must be sorted into different groups in a particular order. In clutter, the order constraints could not be easily satisfied since some objects occlude other objects so the occluded ones are not directly accessible to the robots. Those objects occluding others need to be moved more than once to make the occluded objects accessible. Such rearrangement problems fall into the class of nonmonotone rearrangement problems which are computationally intractable. While the nonmonotone problems with order constraints are harder, involving with multiple robots requires another computation for task allocation.

In this work, we aim to develop a fast method, albeit suboptimally, for the multi-robot coordination for ordered sorting in clutter. The proposed method finds a sequence of objects to be sorted using a search such that the order constraint in each group is satisfied. The search can solve nonmonotone instances that require temporal relocation of some objects to access the next object to be sorted. Once a complete sorting sequence is found, the objects in the sequence are assigned to multiple mobile manipulators using a greedy task allocation method. We develop four versions of the method with different search strategies. In the experiments, we show that our method can find a sorting sequence quickly (e.g., 4.6\,sec with 20 objects sorted into five groups) even though the solved instances include hard nonmonotone ones. The extensive tests and the experiments in simulation show the ability of the method to solve the real-world sorting problem using multiple mobile manipulators.

\end{abstract}

\section{Introduction}
Mobile manipulators have been receiving much attention due to their ability to move and manipulate simultaneously. We see many promising applications such as logistics, manufacturing, and services. There even have been several products ready to be commercialized like Stretch from Boston Dynamics~\cite{stretch_boston} and another Stretch from Hello Robot~\cite{stretch_hello}. 

\begin{figure}
    \centering
    \includegraphics[width=0.5\textwidth]{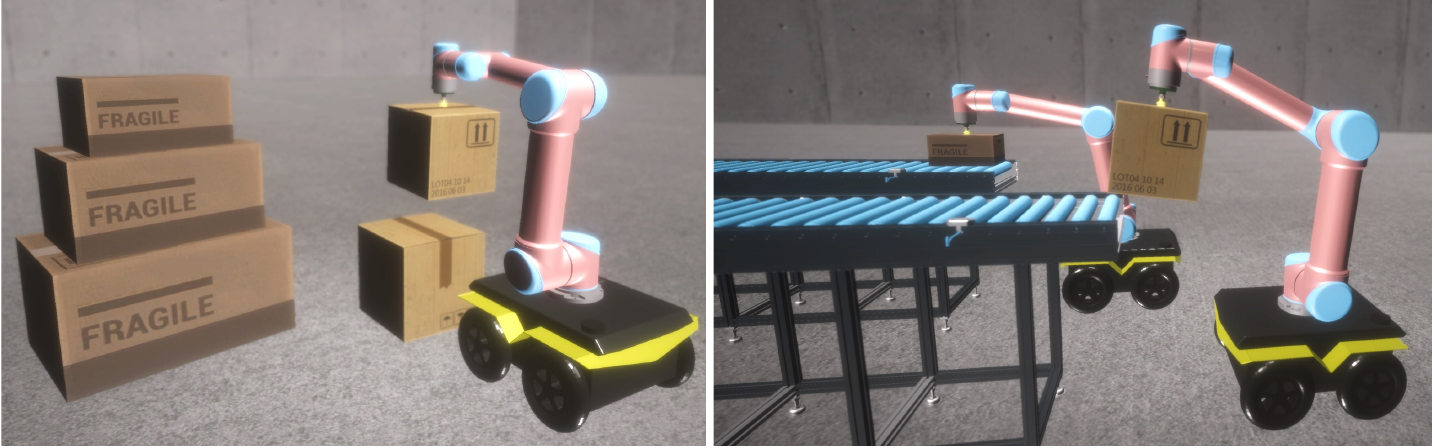}
    \caption{Examples of tasks that should consider the order and type of objects to be manipulated. (Left) Boxes that need to be sorted into groups while the stacking order of objects is important. (Right) Parts or packages that need to be sent to the different destinations in a specific order.}
    \label{fig:example}\vspace{-15pt}
\end{figure}

When we are to manipulate multiple objects, it might be essential to consider what to manipulate first. If we unload a truck or depalletize, the objects (e.g., boxes, totes) to be moved would have a specific order. In general, it is a common practice not to pile up heavy containers on top of small, lighter ones for safety. In a fulfillment center or manufacturing process, products might need to be sent to the next destination in a specified order to meet process requirements. Some items should be delivered in a first-come first-served fashion. At home, we often arrange plates such that smaller plates are piled on larger plates. Along with the need to comply with such order constraints, object sorting problems require groupings of objects according to the characteristics of the objects. Fig.~\ref{fig:example} shows illustrative examples where ordered sorting is necessary.

We aim to solve the ordered sorting problem in clutter using a team of robots. In a dense clutter, some objects need to be relocated more than once (i.e., nonmonotone problem~\cite{stilman2008planning}) if the next object to be sorted is blocked by other objects. This class of problem is more challenging than the multi-robot object rearrangement problem, which is known to be \nphard~\cite{ahn2022coordination} even without the order and group specifications.

We propose a search-based method to find the sequence of objects to be sorted while satisfying the order constraint.  We develop four versions of the method based on different search strategies: Breadth-First Search (BFS), Depth-First Search (DFS), Best-First Search, and A$^*$ Search. Once a sequence is found, we employ a greedy allocation method to assign the objects in the sequence to multiple robots in order to parallelize the sorting task. We show formal analyses of the completeness and optimality and extensive tests to compare the performance of the four versions. We run experiments on the best-performing methods out of the four in a dynamic simulation with multiple mobile manipulators. 




The following are the contribution of this work:
\begin{itemize}
    \item We propose a method to solve the ordered sorting problem in clutter using a team of robots, which is the first attempt in the literature. The method can satisfy the order constraint and handle both the monotone and nonmonotone instances of the problem. 
    \item We provide formal analyses of the proposed method such as completeness, optimality, and time complexity.
    \item We show extensive test results and comparisons of the method. We also test the method using multiple mobile manipulators in dynamic simulation to see how the method works effectively with physical robots.
\end{itemize}

\section{Related Work}

We study the problem of sorting objects in clutter using a team of robots where the order of sorting objects must be satisfied. 
There have been a line of research on multi-robot object manipulation, yet none of them directly tackles the same problem that we are interested in. 

The problem of coordinating multiple robots to transport a common, usually large, object has been solved by various approaches~\cite{feng2020overview,yamashita2003motion,mas2012object} such as path planning and formation control. While these works consider object manipulation using coordinated mobile robots, they focus on transporting a large object together but not sorting objects. 

There are some studies on multi-robot sorting~\cite{yamashita1998cooperative,maneewarn2003sorting}. \cite{yamashita1998cooperative} proposes a trajectory planning method for two robots using a common tool (i.e., stick or string) to sort objects into two groups. \cite{maneewarn2003sorting} also proposes a method to generate a path using a Voronoi diagram where two robots tied to a rope follow the path to separate objects. These works focus on the use of a tool for coordinated sorting without any consideration on the order of sorting. Moreover, they assume that all objects are directly accessible so no nonmonotone instance occurs.

One of the closely related work to ours is \cite{tang2020computing}, which proposes a method to remove multiple objects in a bounded space. The robots have to navigate through a narrow entrance. This work aims to set an order of removal from possible accessibility dependencies between objects in order for them to be reachable, or at least, easier to be reached. Although the method finds a sequence of object removal efficiently even in clutter, it neither assumes removal priorities of objects nor grouping requirements, so it is less constrained. 

Another close work is \cite{ota2009rearrangement}, which considers object rearrangement within a confined space. The proposed method finds a path of a single robot to rearrange (less than ten) objects to achieve a final configuration of the objects. The problem does not directly concern the ordered sorting. Nevertheless, the method can be used for our sorting problem if the final configuration requires the objects to be grouped in a specific order. However, the method generates a path of a single robot, which is not straightforward to distribute the single-robot path to multiple robots. 

Despite these previous efforts, we are not able to find a work that considers the same problem as ours: coordination of multiple robots to sort multiple objects in clutter (with possible occlusions) by given types and priorities. Thus, we aim to fill the gap by proposing an efficient method for rearrangement planning and multi-robot task allocation.

\section{Problem Description}
\label{sec:prob}

\subsection{Assumptions}
\label{sec:assumption}

We have several assumptions to focus on the ordered sorting of cluttered objects: (i) One robot can manipulate one object at a time, and an object needs only one robot to manipulate it. (ii) At least one object must be accessible to one of the robots. (iii) The robots do not use nonprehensile actions. (iv) The environment is known (e.g., geometry of objects, where to sort and temporarily place objects). (v) The number of robots is far lower than the number of objects.

\subsection{Problem formulation}

\begin{figure}[t]
    \centering
   \begin{subfigure}{0.24\textwidth}
   \centering
	\includegraphics[width=0.76\textwidth]{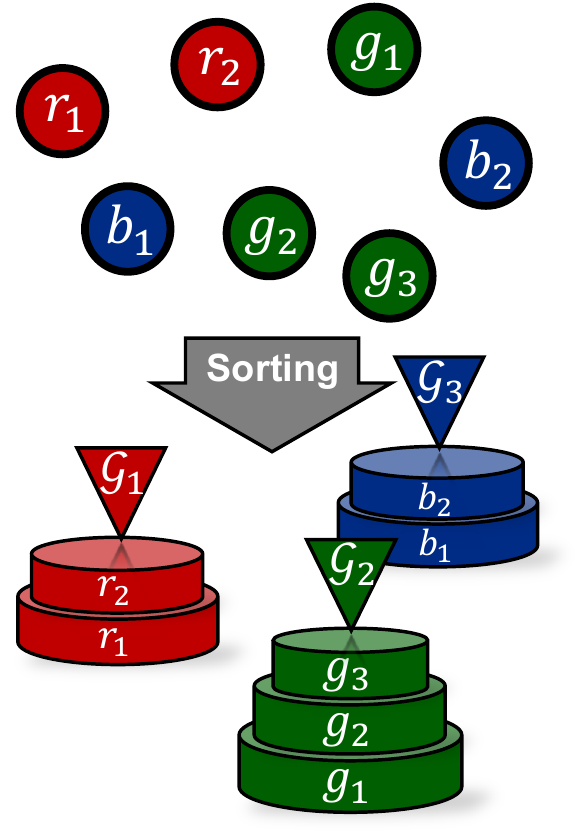}
	\caption{Sorting with order constraints}
    \label{fig:prob_ex_a}
  \end{subfigure}\quad
  \begin{subfigure}{0.19\textwidth}
	\includegraphics[width=0.99\textwidth]{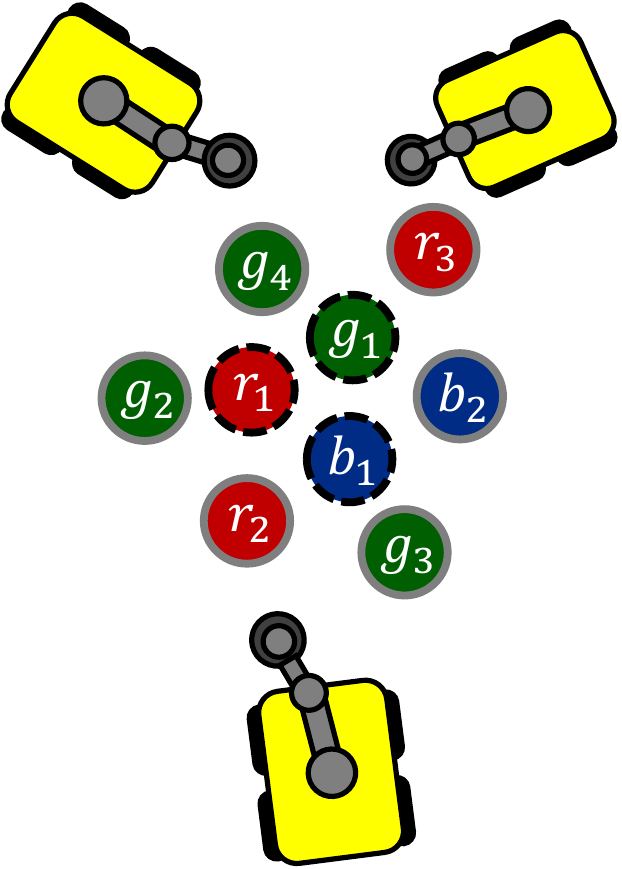}
	\caption{A nonmonotone instance}
    \label{fig:prob_ex_b}
  \end{subfigure}
  \caption{Examples of problem instances. (a) Seven objects are sorted into three groups $\mathcal{G}_1, \cdots, \mathcal{G}_3$. The objects have specific orders to be sorted. (b) In dense clutter, robots would not access all objects directly. The object that must be sorted in the beginning (i.e., $r_1, g_1, b_1$) are occluded.}
  \label{fig:prob_ex}\vspace{-15pt} 
\end{figure}

In a workspace $\mathcal{W}$, the goal is to sort $N$ objects in $\mathcal{O}$ into $K$ groups using a team of $M$ robots in $\mathcal{R}$. The objects have a specific order to be sorted. A group $\mathcal{G}_i$ for $i = 1, \cdots, K$ is a tuple of objects where $|\mathcal{G}_i| > 0$ and $\sum_i^K |\mathcal{G}_i| = |\mathcal{O}| = N$. Throughout the paper, we will use different object colors, such as \textit{red}, \textit{green}, \textit{blue}, and so on, to better distinguish objects in different groups. Without loss of generality, we assume that the indices of objects in the groups are in the ascending order. In Fig.~\ref{fig:prob_ex_a}, we have $\mathcal{G}_1 = (r_1, r_2)$,  $\mathcal{G}_2 = (g_1, g_2, g_3)$, and $\mathcal{G}_3 = (b_1, b_2)$. In $\mathcal{G}_1$, $r_1$ must be sorted before $r_2$. Without occlusions between objects, sorting an object belonging to one group is independent from other groups. On the other hand, the objects in different groups would be interrelated if an object in one group is blocked by other objects in different groups. 

The problem consists of two challenging subproblems. The first subproblem is the ordered sorting problem in which occlusions between objects could prevent robots from complying with the order constraint. This subproblem mandates solving the manipulation planning among movable obstacles (MAMO) problem, which is proven to be \nphard~\cite{stilman2008planning}. Furthermore, our problem needs to consider the order constraint. The second subproblem is the multi-robot task allocation (MRTA), determining which robots sort which objects. Since we assume $M < N$, the allocation should consider an extended time frame, which falls into the strongly \nphard ST-SR-TA problem (\textbf{S}ingle-\textbf{T}ask robots, \textbf{S}ingle-\textbf{R}obot tasks, \textbf{T}ime-extended \textbf{A}ssignment) according to the MRTA taxonomy~\cite{gerkey2004formal}. As the entire problem is computationally intractable, developing an optimal method is not practically useful. Thus, we solve the subproblems for faster planning and the sorting task separately.

We aim to find a sequence of objects to be sorted, or a tuple $\mathcal{O}_S$, with the minimum length. Then we repeatedly assign the robots to the objects in $\mathcal{O}_S$ until all objects are sorted.
In the example in Fig.~\ref{fig:prob_ex_a}, a valid solution is $\mathcal{O}_S = (r_1, g_1, r_2, g_2, b_1, g_3, b_2)$. Another solution $\mathcal{O}_S = (g_1, g_2, g_3, r_1, r_2, b_1, b_2)$ is also valid. Both sequences result in the stacks shown in Fig.~\ref{fig:prob_ex_a}. In these monotone instances, $|\mathcal{O}_S| = N$. If some objects are not accessible unless the occluding objects are relocated, the problem instance is nonmonotone and hence $|\mathcal{O}_S| > N$. As shown in Fig.~\ref{fig:prob_ex_b}, none of the first objects in $\mathcal{G}_{1}, \cdots, \mathcal{G}_3$ is accessible, so one or more of the occluding objects should be relocated. Even in this nonmonotone case, we want to minimize $|\mathcal{O}_S|$ by relocating only the necessary objects. 

\section{Methods}
We propose our method that solves the sorting problem using search and a greedy MRTA method. 

\subsection{Search-based planning}
\label{sec:search}

We develop four versions of the planning method that are based on uninformed search (Breadth-First and Depth-First Search) and informed search (Best-First and $A^*$ Search). They share most procedures shown in Alg.~\ref{alg:sorting}. They have different implementations for the frontier (lines~\ref{line:method_s}--\ref{line:method_e}) and evaluation functions (lines~\ref{line:cost_s}--\ref{line:cost_e}). The priority queue used in the informed search enqueues the element with the minimum value. 
Other specific implementations of functions are explained in Sec.~\ref{sec:utility}.

In Alg.~\ref{alg:sorting}, each node in the search tree represents an object configuration (i.e., the status of $\mathcal{O}$ and $\mathcal{G}_i$). The root represents the initial object configuration with empty depots. Since the root is inserted into the frontier (line~\ref{line:frontier_init}), the while loop (lines~\ref{line:while_s}--\ref{line:while_e}) iterates until all objects are sorted (line~\ref{line:success}) or no object can be manipulated anymore (line~\ref{line:infeasible}).\footnote{In Sec.~\ref{sec:assumption}, we assume that at least one object can be manipulated. Thus, the instances that we consider do not incur the latter infeasible case.}

In the while loop, line~\ref{line:get} chooses the node to expand, which varies depending on the search strategy. The node chosen to expand represents a particular object configuration where some objects are sorted or temporarily sent to arbitrarily chosen free spaces that we refer as buffer locations. Among accessible objects (line~\ref{line:getaccessible}), line~\ref{line:intersection} finds the objects that can be sorted while complying with the order constraint. Suppose that the accessable objects $\mathcal{O}_A = \{g_1, g_3, b_3\}$, the sets of objects in the depots $\mathcal{G}_1 = \{r_1\}$, $\mathcal{G}_2 = \emptyset$, and $\mathcal{G}_3 = \{b_1, b_2\}$. Then the objects to be sorted next are $\mathcal{O}_G = \{r_2, g_1, b_3\}$. Thus, the next objects to allocate to robots are $\mathcal{O}_N = \mathcal{O}_A \cap \mathcal{O}_G = \{g_1, b_3\}$. 

If $\mathcal{O}_N = \emptyset$, all objects in $\mathcal{O}_G$ are not accessible owing to occlusion (so nonmonotone). At least one object in $\mathcal{O}_G$ needs to become accessible to continue. Thus, lines~\ref{line:nonmonotone_s}--\ref{line:nonmonotone_e} find the minimum number of objects (line~\ref{line:shortest}) to be relocated to buffers temporarily. While they are waiting in the buffers, they are included in $\mathcal{O}_A$ so can be sorted in their turn.

If $\mathcal{O}_N \neq \emptyset$ or $\mathcal{O}_N$ becomes nonempty after relocating occluding objects, all objects in $\mathcal{O}_N$ are sorted so one or more depots are filled (line~\ref{line:sort}). After sorting, lines~\ref{line:insert_s}--\ref{line:insert_e} add a new node to the frontier. Once a goal node is found, $\mathcal{O}_S$ contains all objects to be sorted. In nonmonotone instances, it contains some objects sent to buffers temporarily. If the search finishes without returning \textsf{Success}, the complete sorting sequence 
cannot be found (line~\ref{line:failure}).\footnote{In Sec.~\ref{sec:analysis}, we show that Alg.~\ref{alg:sorting} is complete so never returns \textsf{Failure} (Theorem 4.2).}

\begin{algorithm}
\caption{{\scshape SortObj}} \label{alg:sorting}
\begin{algorithmic}[1]
{\footnotesize
\floatname{algorithm}{Procedure}
\renewcommand{\algorithmicrequire}{\textbf{Input:} }
\renewcommand{\algorithmicensure}{\textbf{Output: }}
\REQUIRE Objects $\mathcal{O}$, $K$, robots $\mathcal{R}$, workspace $\mathcal{W}$, $method$
\ENSURE Result

\IF{$method = $ \textsf{BFS}} \label{line:method_s}
    \STATE Initialize a queue F as the frontier 
\ELSIF{$method = $ \textsf{DFS}}
    \STATE Initialize a stack F as the frontier 
\ELSIF{$method = $ \textsf{BestFirst} \OR $method = $ \textsf{AStar}}
    \STATE Initialize a min priority queue F as the frontier
\ENDIF \label{line:method_e}
\STATE Initialize $K$ stacks: $\mathcal{G}_1, \cdots, \mathcal{G}_K$ \label{line:group}
\STATE Let $v$ be the node representing the initial configuration of $\mathcal{O}$
\IF{$method = $ \textsf{BestFirst}}
    \STATE $v$.value = 0 \label{line:best_init}
\ENDIF
\STATE F\textsc{.Enqueue}($v$) \codecomment{Implemented as \textsc{Push} for \textsf{DFS}} \label{line:frontier_init}
\WHILE{F is not empty} \label{line:while_s}
    \STATE $v=$ F\textsc{.Dequeue}() \codecomment{Implemented as \textsc{Pop} for \textsf{DFS}} \label{line:get}
    \IF{$\sum_i^K |\mathcal{G}_i| = |\mathcal{O}|$}
        \RETURN \textsf{Success} \label{line:success}
    \ENDIF
    \STATE $\mathcal{O}_A =$ \textsc{GetAccessibleObjs}$(\mathcal{O}, \mathcal{R}, \mathcal{W})$ \codecomment{Get accessible objects from robots} \label{line:getaccessible}
    \IF{$\mathcal{O}_A = \emptyset$}
        \RETURN \textsf{Infeasible} \label{line:infeasible} \codecomment{No object can be manipulated}
    \ENDIF
    \STATE $\mathcal{O}_G =$ \textsc{GetNextObjs}$(\mathcal{G}_1, \cdots, \mathcal{G}_K)$ \codecomment{Get the set of the objects to be sorted next}\label{line:getnext}
    \STATE $\mathcal{O}_N = \mathcal{O}_A \cap \mathcal{O}_G$ \label{line:intersection} 
    \IF{$\mathcal{O}_N = \emptyset$} \label{line:nonmonotone_s}
        \FOR{each object $o$ in $\mathcal{O}_G$} \label{line:for_s}
            \STATE $\mathcal{O}_R=$ \textsc{RelocObjs}$(\mathcal{O}, o, \mathcal{R}, \mathcal{W})$\codecomment{Find objects to be relocated to make $o$ accessible (Sec.~\ref{sec:utility})}
            \STATE Keep $\mathcal{O}_R$ shortest \label{line:shortest} \codecomment{Send the least objects to buffer}
        \ENDFOR \label{line:for_e}
        \STATE Relocate objects in $\mathcal{O}_R$ to buffers \label{line:buffer} \codecomment{Nonmonotone}
        \STATE $\mathcal{O}_A =$ \textsc{GetAccessibleObjs}$(\mathcal{O}, \mathcal{R}, \mathcal{W})$
        \STATE $\mathcal{O}_N = \mathcal{O}_A \cap \mathcal{O}_G$
    \ENDIF \label{line:nonmonotone_e}
    \FOR{each object $o$ in $\mathcal{O}_N$} \label{line:sort_s}
        \STATE Get the group index $i$ of $o$ 
        \STATE $\mathcal{G}_i$.\textsc{Push}$(o)$ \label{line:sort} \codecomment{Sort $o$ to its group}
    \ENDFOR 
    \STATE Let $v$ be the node representing the current config. of $\mathcal{O}$  \label{line:insert_s}
    \IF{$method = $ \textsf{BestFirst}}\label{line:cost_s}
        \STATE $v$.value = $h(v)$ \codecomment{Detailed in Sec.~\ref{sec:cost}} 
    \ELSIF{$method = $ \textsf{AStar}}
        \STATE $v$.value = $h(v) + g(v)$\codecomment{Detailed in Sec.~\ref{sec:cost}}
    \ENDIF\label{line:cost_e}
    \STATE F\textsc{.Enqueue}($v$) \label{line:insert_e}
\ENDWHILE \label{line:while_e}
\RETURN \textsf{Failure} \codecomment{Search fails if the frontier is empty} \label{line:failure}
}
\end{algorithmic}
\end{algorithm}
\subsubsection{Heuristics for informed search}
\label{sec:cost}

Let $f(v)$, $h(v)$, and $g(v)$ be the evaluation, heuristic, and path cost function for node $v$, respectively. 
Since our goal is to minimize $\mathcal{O}_S$, $h(v)$ is the number of unsorted objects that are not yet in $\mathcal{G}$. In other words, $h(v) = N_R = N - \sum_{i=1}^K |\mathcal{G}_i|$. If monotone, the robots need to manipulate only the remaining $N_R$ objects. Otherwise, if nonmonotone, the robots manipulate more than $N_R$ objects since some objects are sent to buffers. The actual number of manipulated objects is always less than or equal to $h(v)$, which is an admissible heuristic that never overestimates the number of manipulated objects. 
The path cost function $g(v)$ is simply the number of manipulated objects from the root to node $v$. 

Due to the order constraint, a robot transporting an object to a depot should wait until a currently working robot at the depot leaves. Frequent waiting decreases the throughput of sorting. Also, there may be higher chances of crashes or deadlocks if many robots wait at a depot. We deal with this problem by defining a secondary criterion for the search for the nodes with the same $f(v)$. We penalize $f(v)$ if objects belonging to the same group are in the sorting sequence consecutively. Given $M$ robots, we consider the most recent $M$ objects in the current $\mathcal{O}_S$ to compute the penalty. The window size could vary depending on the environment.

\subsubsection{Utilities}
\label{sec:utility}
Function \textsc{GetAccessibleObjs} finds the objects that are accessible to robots. 
We employ a polynomial-time method proposed in~\cite{lee2019efficient} to determine the accessibility of an object without performing motion planning. The method needs to know only the sizes of an object and the end-effector to find a collision-free direction that the end-effector can approach to grasp the object. Since this method does not consider the kinematic constraints of the whole robot, an object that is determined to be accessible could turn out to be inaccessible in motion planning. Nevertheless, it dramatically decreases the planning time by removing the computation-intensive motion planning as demonstrated in~\cite{nam2020fast,nam2021fast,ahn2021integrated}. 

Next, \textsc{GetNextObjs} finds the next objects to be sorted in each group $\mathcal{G}_1, \cdots, \mathcal{G}_K$. The input is the groups implemented by stacks. \textsc{Top} operation for all $K$ groups tells the lastly sorted objects in each group. Then the next object to be sorted in each group can be found. Last, \textsc{RelocObjs} finds the objects to be relocated to make an inaccessible object ($o$ in the input) accessible. If the objects to be sorted found by \textsc{GetNextObjs} are not accessible, this function tells what to relocate to buffers. We employ a method proposed in~\cite{nam2021fast} that runs in polynomial time in the number of objects. 

\subsection{Greedy task allocation}
\label{sec:allocation}
With $N > M$, an assignment between $\mathcal{O}_S$ and $\mathcal{R}$ can be computed by repeatedly assigning a few objects to a few idle robots in a batch. However, this batch allocation requires some robots to wait until a sufficient number of robots become idle. Thus, we implement an online greedy task allocation algorithm assigning each object in $\mathcal{O}_S$ to each idle robot that can access the object. 
If more than one robot can access the object, the closest robot is assigned.

\subsection{Analysis of the algorithms}
\label{sec:analysis}

\lem \textbf{4.1.} The heuristics $h(v)$ in Alg.~\ref{alg:sorting} is admissible.

\noindent We omit the proof due to the space limit, but the lemma is already proved informally in Sec.~\ref{sec:cost}.

\thm \textbf{4.2.} All versions of Alg.~\ref{alg:sorting} are complete.

\pf. 
BFS is known to be complete. A$^*$ Search is complete with an admissible heuristic.\footnote{All the well-known proofs are given in~\cite{russell2002artificial}.} DFS and Best-First Search are complete if the state space is finite. In our sorting problem, the flow of objects is one-way: from the clutter to depots directly or via buffers. They never move back to the clutter from the depots or buffers. Sorted objects in the depots do not move. Objects in buffers are accessible, so they do not move inside the buffers. Thus, our problem is with a finite state space. 
\qed

\thm \textbf{4.3.} Alg.~\ref{alg:sorting} with BFS and A$^*$ are optimal.

\pf. 
BFS is optimal if the path cost is a non-decreasing function of the depth of the node~\cite{russell2002artificial}. As described in Sec.~\ref{sec:cost}, $g(v)$ is the number of manipulated objects until $v$, which must increase as the search depth increases. A$^*$ is optimal with an admissible heuristic. Thus, both versions of Alg.~\ref{alg:sorting} are optimal in the length of $\mathcal{O}_S$. 
\qed

\propo \textbf{4.4.} Each generation of a search node in Alg.~\ref{alg:sorting} runs in polynomial time. 

\pf. 
All the search methods run in exponential time in the depth of the search tree $O(b^d)$ where $b$ is the largest branching factor and $d$ is the depth. In the worst case, both quantities can be up to $N$. Although the entire search could take exponential time, each node generation is done efficiently (lines~\ref{line:while_s}--\ref{line:while_e}). Our implementation of \textsc{GetAccessibleObjs} has $O(N^3)$ time complexity. Inside \textsc{GetAccessibleObjs}, the accessibility check for each object takes $O(N^2)$~\cite{lee2019efficient}, which repeats up to $N$ times. \textsc{GetNextObjs} runs the constant-time top operation of stacks $K$ times. Thus, the function runs in $O(K)$.
The most excessive computation could occur in lines~\ref{line:for_s}--\ref{line:for_e}, where \textsc{RelocObjs} repeats $|\mathcal{O}_G|$ times. The quantity $|\mathcal{O}_G|$ is bounded by the number of groups $K$ as \textsc{GetNextObjs} finds one object to be sorted in each group. In \cite{nam2021fast}, \textsc{RelocObjs} is shown to be with $O(N^4)$. Thus, the time complexity of the entire for loop is $O(N^4 K)$, which dominates all the operations in the while loop.
\qed

Even though the algorithms do not have polynomial time complexity, DFS and Best-First Search find a solution quickly, which will be shown in experiments (Sec.~\ref{sec:exp}). We also show that the solution quality of DFS and Best-First are the same as the optimal solution from A$^*$ even though the two methods are not guaranteed to find an optimal solution.

\section{Experiments}
\label{sec:exp}

\begin{figure*}
    \centering
    \includegraphics[width=0.99\textwidth]{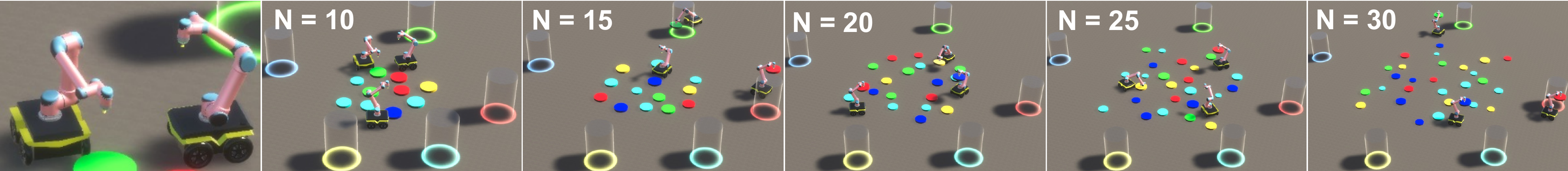}
    \caption{Example scenes of the experiments in simulation for $K = 5$ (objects with five colors) and varying $N$. The left picture shows the mobile manipulators (UR5 and Jackal) that manipulate objects using a suction gripper.}
    \label{fig:unity}\vspace{-5pt}
\end{figure*}

The first set of experiments evaluates the performance of Alg.~\ref{alg:sorting} with a vast amount of random instances. Next, we run another set in a dynamic simulation using the Unity-ROS integration (Fig.~\ref{fig:unity})~\cite{unity}. All experiments are done with the system with AMD 5800X 3.8GHz/32G RAM, Python 3.8.

\subsection{Algorithm tests} 
\label{sec:test}

We increase $N$ from 10 to 30 at intervals of 5 and have 1, 3, and 5 for $K$. For each $(N, K)$, we generate 20 instances where object locations are sampled uniformly at random. The number of objects that belong to the same group is randomly sampled while zero object in a group is not allowed. The object shape is disc whose diameter is randomly sampled around 30\,cm. Then, we use the same set of 300 instances for testing the four versions for a fair comparison. Some instances are visualized using the simulator (Fig.~\ref{fig:unity}). 

We measure the objective value, the length of $\mathcal{O}_S$, which indicates the number of objects that the robots should manipulate until the end. Another important metric is the task planning time measuring how long Alg.~\ref{alg:sorting} takes to generate $\mathcal{O}_S$. 
We do not have the notion of robots in this algorithm test. Thus, the task planning time does not include the time for motion planning or task allocation. 

Other metrics include the success rate of task planning given a time limit (30\,sec). Although the proposed algorithm is shown to be complete, we want to achieve fast planning. Thus, we set the tight limit even for the most challenging instance with $(N, K) = (30, 1)$. We also calculate the repetitiveness of a sorting sequence $\mathcal{O}_S$ to see the effect of adopting the secondary criterion for breaking ties (presented in Sec.~\ref{sec:cost}). This metric measures how many objects belonging to the same group are listed in $\mathcal{O}_S$ in a row. A higher value means robots often use the same depot in series.

Although having only one group ($K = 1$) makes the problem merely stacking objects but not sorting, we test it since the problem becomes more difficult. The function \textsc{GetNextObjs} in line~\ref{line:getnext} of Alg.~\ref{alg:sorting} gives at most $|\mathcal{O}_G| = K$ objects. The chance to have fewer objects in $\mathcal{O}_N = \mathcal{O}_A \cap \mathcal{O}_G$ increases with a smaller $K$. A larger $N$ incurs more occlusions between objects so $|\mathcal{O}_A|$ is likely to decrease, resulting in small $|\mathcal{O}_N|$. If $\mathcal{O}_N = \emptyset$, we need to solve the nonmonotone problem. As an example, we show the ratio of nonmonotone instances out of 20 random instances in Table~\ref{tab:nonmonotone}. As $K$ decreases and $N$ increases, a significant portion of test instances fall to the nonmonotone class. We show the length of $\mathcal{O}_S$ in Table~\ref{tab:len}. Since $N=10$ is the only case where all the four versions can finish without excessive running time, we show the result for $N=10$. As expected, the length $|\mathcal{O}_S|$ increases as $K$ decreases. This can be justified by the fact that a small $K$ leads to more nonmonotone instances appear, which require additional moves to buffers.

\begin{table}
    \caption{Measures that help understand the level of difficulty of test instances. (a) More nonmonotone instances as $N$ increases and $K$ decreases. (b) The more nonmonotone instances the longer sequence.}
    \begin{subtable}{0.24\textwidth}
        \centering
        \scalebox{0.9}{%
        \begin{tabular}{|c|c||c|c|c|}
        \hline
         \multicolumn{2}{|c||}{$K$} & \textsc{$1$} & \textsc{$3$} & \textsc{$5$} \\
        \hline
        \multirow{5}{*}{$N$} & $10$ & 0.5 & 0.15 & 0 \\
        \cline{2-5}
        & $15$ & 0.8 & 0.2 & 0 \\
        \cline{2-5}
        & $20$ & 1.0 & 0.35 & 0.05 \\
        \cline{2-5}
        & $25$ & 0.95 & 0.3 & 0.15 \\
        \cline{2-5}
        & $30$ & 0.95 & 0.5 & 0 \\
        \hline
        \end{tabular}}
        \caption{Ratio of nonmonotone instances}
        \label{tab:nonmonotone}
    \end{subtable}
    \begin{subtable}{0.24\textwidth}
        \centering
        \scalebox{0.9}{%
        \begin{tabular}{|c|c||c|c|c|}
        \hline
         \multicolumn{2}{|c||}{$K$} & \textsc{$1$} & \textsc{$3$} & \textsc{$5$} \\
        \hline
        \multirow{5}{*}{$N$} & $10$ & 10.85 & 10.15 & 10.0 \\
        \cline{2-5}
        & $15$ & 16.75 & 15.3 & 15.0 \\
        \cline{2-5}
        & $20$ & 22.75 & 20.5 & 20.05 \\
        \cline{2-5}
        & $25$ & 27.65 & 25.55 & 25.2 \\
        \cline{2-5}
        & $30$ & 33.4 & 30.85 & 30.0 \\
        \hline
        \end{tabular}}
        \caption{The mean length of $\mathcal{O}_S$}
        \label{tab:len}
     \end{subtable}
     \label{tab:difficulty}
     \vspace{-10pt}
\end{table}

\begin{figure}
    \centering
   \begin{subfigure}{0.22\textwidth}
   \centering
    \includegraphics[width=\textwidth]{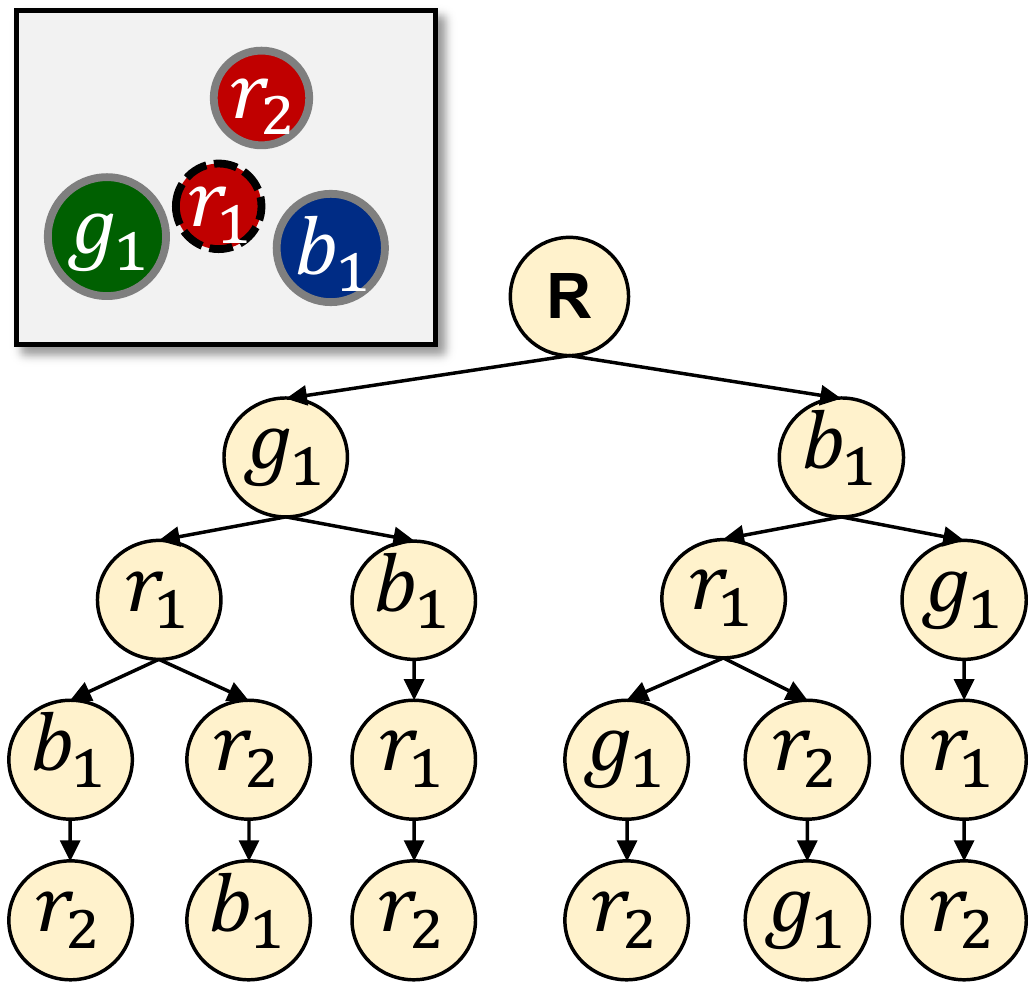}
	\caption{Monotone case}
    \label{fig:length1}
  \end{subfigure}
  \begin{subfigure}{0.42\textwidth}
  	\includegraphics[width=\textwidth]{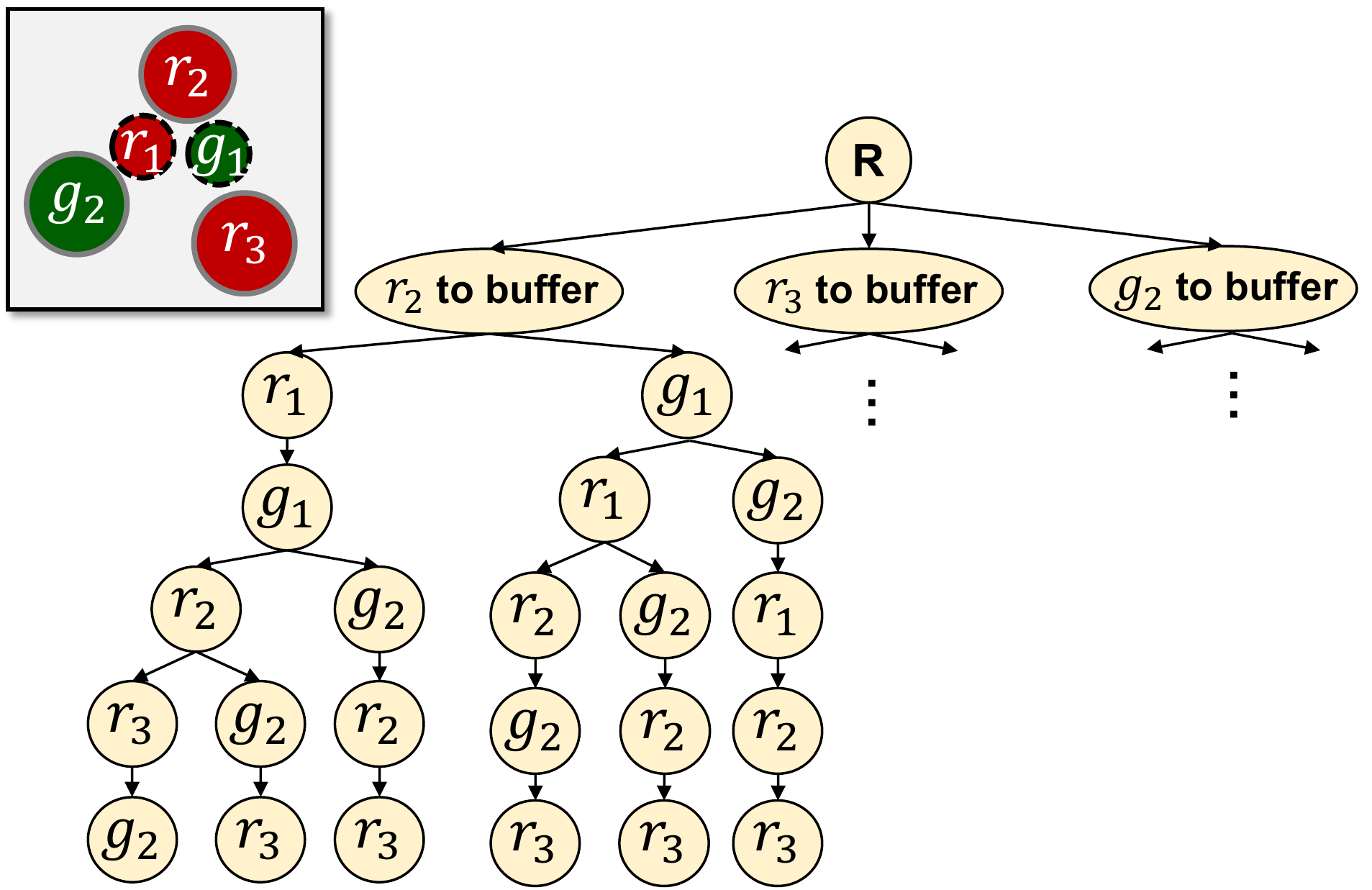}
	\caption{Nonmonotone case}
    \label{fig:length2}
  \end{subfigure}
  \caption{Example complete search trees where the leaves have the same depth}
  \label{fig:length}
  \vspace{-15pt} 
\end{figure}

Interestingly, all four versions have the same $|\mathcal{O}_S|$ if they run the same instance. Fig.~\ref{fig:length}) shows two complete search trees for a toy monotone and a nonmonotone instance in the gray boxes. They have inaccessible objects denoted by dotted outlines. With a slight abuse of notation, we inscribe the object to be sorted in the search node except for the root node and the node where an object is sent to buffers. All leaf nodes have the same depth, which all represent the goal states. Depending on the search strategy, one of the leaf nodes is reached via possibly different paths. In any case, sorting is completed with the minimum $|\mathcal{O}_S|$, which is possible owing to the way Alg.~\ref{alg:sorting} generates nodes. In each iteration (lines~\ref{line:while_s}--\ref{line:while_e}), objects are manipulated only necessarily resulting in $|\mathcal{O}_S|= N$ obviously in the monotone case. In the nonmonotone case, some objects should be sent to buffers. In Fig.~\ref{fig:length2}, we omit part of the tree owing to the space limit but the subtrees from the children of the root have the same height. However, some instances would have situations where one or more of the branches have different numbers of objects sent to buffers so the depths of the leaf nodes are not identical. Nevertheless, the four versions produce the same $|\mathcal{O}_S|$ for the same instance in our tests. Probably many instances have the trees like those shown in Fig.~\ref{fig:length}. Otherwise, all search methods seem to reach the goal node at the shallowest level.

\begin{figure}
    \centering
   \begin{subfigure}{0.28\textwidth}
   \centering
    \includegraphics[width=\textwidth]{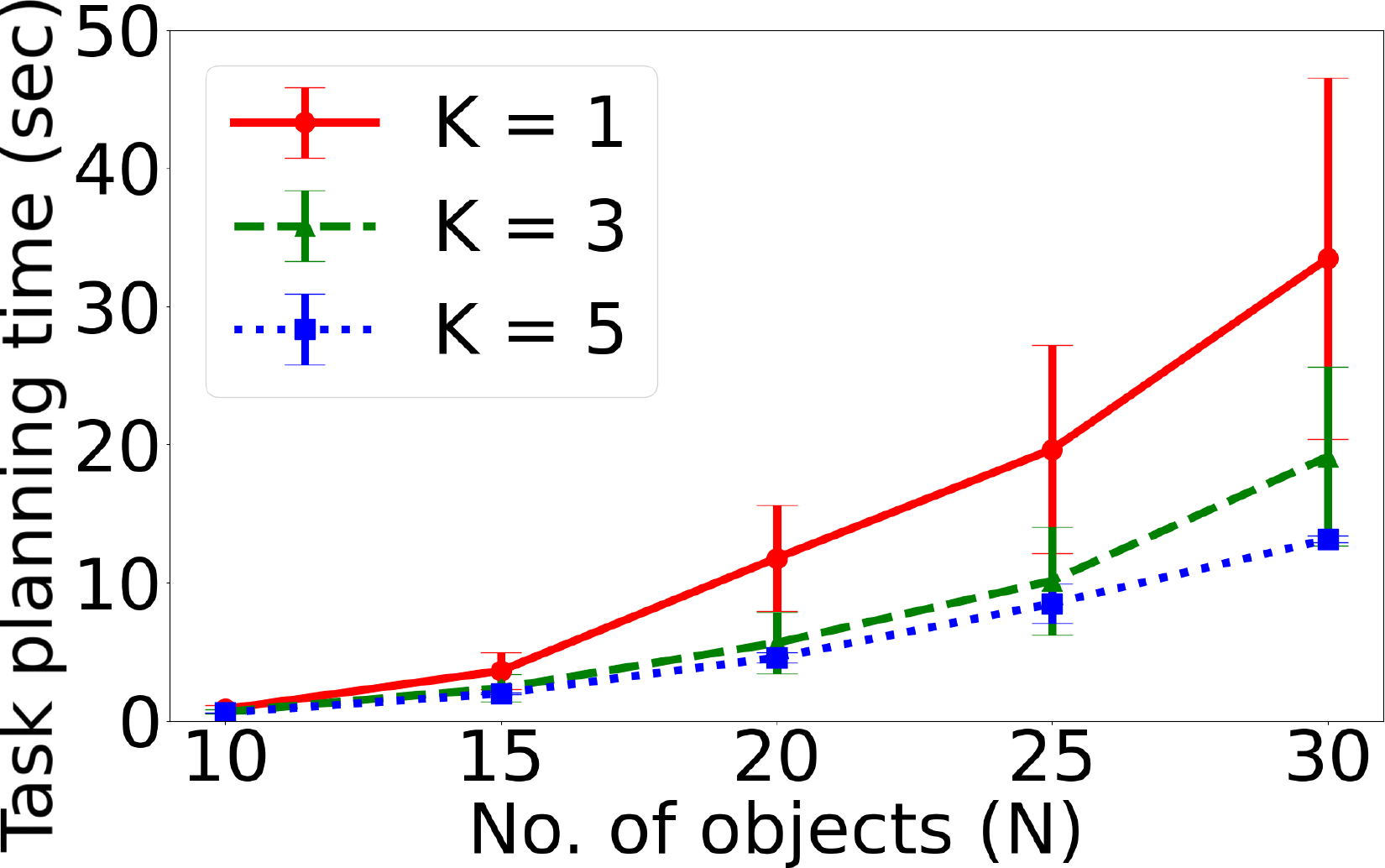}
	\caption{Best-First Search}
    \label{fig:task1}
  \end{subfigure}
  \begin{subfigure}{0.28\textwidth}
  	\includegraphics[width=\textwidth]{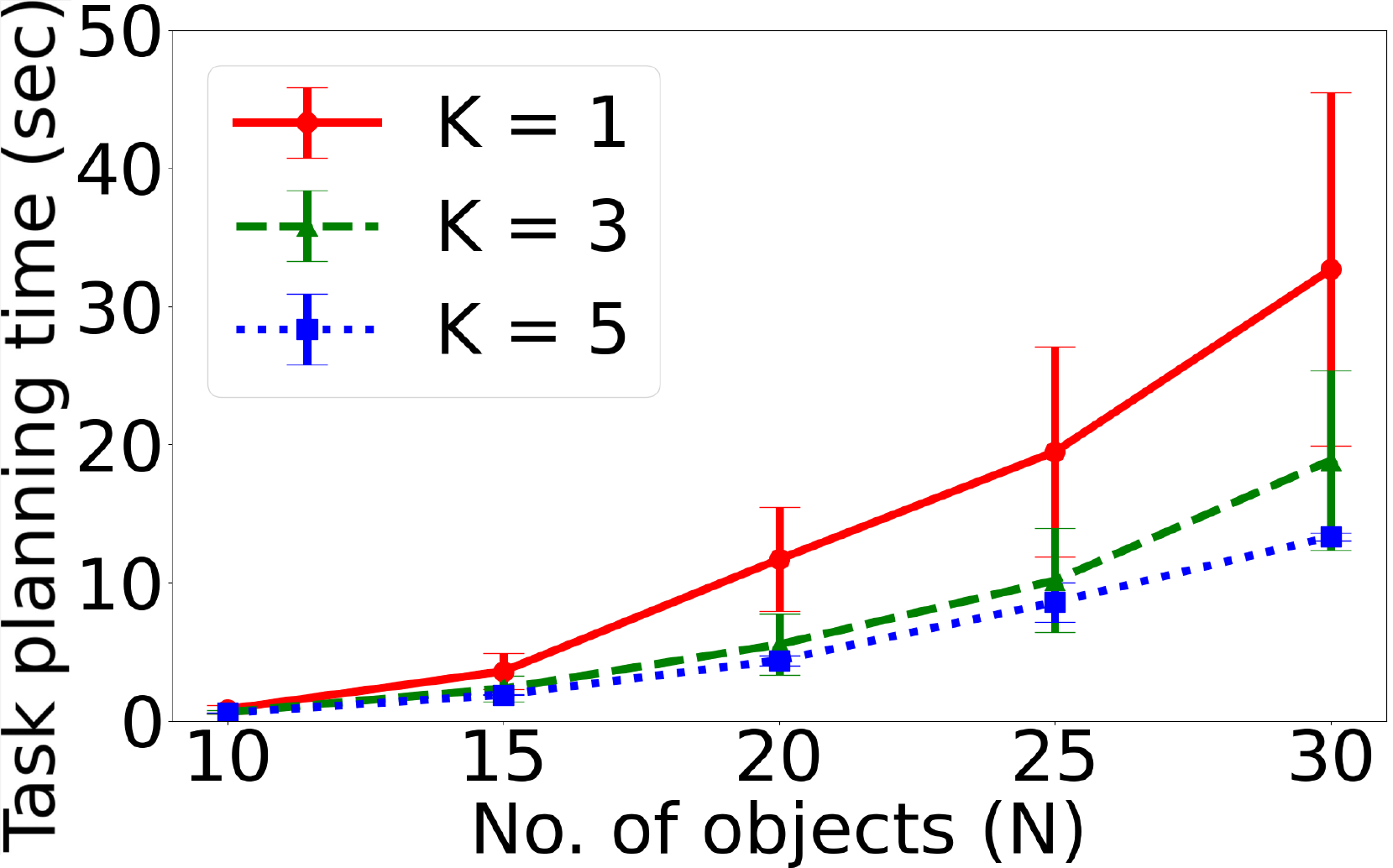}
	\caption{Depth-First Search}
    \label{fig:task2}
  \end{subfigure}
  \caption{The computation time of Alg.~\ref{alg:sorting} to generate a sorting sequence. Since BFS and A$^*$ are prohibitively slow with $N > 10$, we only show the results of two fast versions of Alg.~\ref{alg:sorting}.}
  \label{fig:task}
  \vspace{-10pt}
\end{figure}

The computation time of the four versions is shown in Fig.~\ref{fig:task} and Table~\ref{tab:task}. BFS and A$^*$ show prohibitively long time even for moderately-sized instances so we do not test them beyond $N = 10$ so show the result only in Table~\ref{tab:task2}. Best-First and DFS are the fastest. As discussed above, any leaf node in the tree is a goal node. DFS directly goes down deep so reaches the leftmost leaf quickly. The heuristic function of Best-First Search is designed to expand the node with a lesser number of unsorted objects. Thus, it is likely to choose the node which is at the greater depth rather than the nodes at the same level. Since both tend to go down in the tree, an increase in the branching factor (closely related to $K$) does not significantly affect the search time. Rather, an increase of $K$ allows more objects in $\mathcal{O}_G$ so causes a less number of nonmonotone instances (Table~\ref{tab:nonmonotone}). Thus, the running time decreases as $K$ increases. It is clear that the running time increases with larger $N$ as shown in Fig.~\ref{fig:task}.  

Since BFS explores all nodes at the current depth, a higher branching factor (with a larger $K$) increases the running time significantly (Table~\ref{tab:task2}). In A$^*$, the evaluation function adds the number of unsorted objects ($h(v)$) and the number of sorted objects so far. For those trees with leaves with an identical depth, $h(v)$ is the same for all nodes in the same level. Thus, the search behaves like BFS. Nevertheless, A$^*$ is faster because it can explore a deeper level if some leaf nodes have smaller depths so the search goes down.

\begin{table}
    \caption{The average running time and standard deviation of Alg.~\ref{alg:sorting} in seconds from 20 random instances for each $(N, K)$. BFS and A$^*$ run slow with large $N$ so the result with $N = 10$ is shown. }
    \begin{subtable}{0.49\textwidth}
        \centering
        \scalebox{0.82}{%
        \begin{tabular}{|c|c|c||c|c|c|c|c|c|}
        \hline
         \multicolumn{3}{|c||}{Method} & \multicolumn{3}{c|}{Best-First} & \multicolumn{3}{c|}{Depth-First} \\
        \hline
         \multicolumn{3}{|c||}{$K$} & \textsc{$1$} & \textsc{$3$} & \textsc{$5$} & \textsc{$1$} &  \textsc{$3$} & \textsc{$5$} \\
        \hline
         \multirow{10}{*}{$N$} & \multirow{2}{*}{$10$} & Mean & 0.8498 & 0.6608 & 0.5957 & 0.8419 & 0.6329 & 0.5714 \\
         \cline{3-9} && SD & 0.3147 & 0.1554 & 0.0130 & 0.3112 & 0.1487 & 0.0131 \\
         \cline{2-9} & \multirow{2}{*}{$15$} & Mean & 3.6190 & 2.3722 & 1.9835 & 3.5848 & 2.3404 & 1.8931 \\
         \cline{3-9} && SD & 1.3131 & 0.9614 & 0.0331 & 1.2942 & 0.9195 & 0.0304 \\
         \cline{2-9} & \multirow{2}{*}{$20$} & Mean & 11.7834 & 5.6461 & 4.5822 & 11.6923 & 5.5343 & 4.3815 \\
         \cline{3-9} && SD & 3.3879 & 2.2231 & 0.3797 & 3.7569 & 2.2278 & 0.3786 \\
         \cline{2-9} & \multirow{2}{*}{$25$} & Mean & 19.6516 & 10.1491 & 8.4923 & 19.4819 & 10.1660 & 8.5880 \\
         \cline{3-9} && SD & 7.5329 & 3.9095 & 1.4130 & 7.5891 & 3.7654 & 1.4156 \\
         \cline{2-9} & \multirow{2}{*}{$30$} & Mean & 33.4812 & 19.1276 & 13.1461 & 32.6828 & 18.8620 & 13.3311 \\
         \cline{3-9} && SD & 13.0828 & 6.4647 & 0.2361 & 12.8003 & 6.4927 & 0.3065 \\
        \hline
        \end{tabular}}
       \caption{Best-First and Depth-First Search}
       \label{tab:task1}
       \vspace{8pt}
    \end{subtable}
    \begin{subtable}{0.49\textwidth}
        \centering
        \scalebox{0.88}{%
        \begin{tabular}{|c||c|c|c|c|c|c|}
        \hline
         Method & \multicolumn{3}{c|}{Breadth-First}  & \multicolumn{3}{c|}{A$^*$} \\
        \hline
         \ $K$ & \textsc{$1$} & \textsc{$3$} & \textsc{$5$} & \textsc{$1$} & \textsc{$3$} & \textsc{$5$} \\
        \hline
        Mean & 0.8349 & 50.7530 & 641.6147 & 0.8392 & 3.0477 & 45.4273 \\
        \hline
        \ STD & 0.3115 & 40.0886 & 441.8589 & 0.3028 & 2.5476 & 31.7290\\
        \hline
        \end{tabular}}
        \caption{Breadth-First and A$^*$ Search}
        \label{tab:task2}
     \end{subtable}
     \label{tab:task}
\end{table}

The success rate is summarized in Table~\ref{tab:success_rate}. As expected, Best-First and DFS show higher success rates even with the tight time limit (30\,sec). As discussed, the instances with $K=1$ do not represent the sorting problem so we sideline them from discussion. With $K > 1$, the search finishes quickly in the most of the instances. With a slightly longer time limit (i.e., 32\,sec), the success rates achieve 100\% for all instances with $K > 1$. BFS and A$^*$ are not adequate for those applications where a lack of responsiveness causes catastrophic failures. 

\begin{table}
\caption{The success rate (\%) of task planning within 30\,sec}
\label{tab:success_rate}
\centering%
\scalebox{0.79}{%
\begin{tabular}{|c|c||c|c|c|c|c|c|c|c|c|c|c|c|}
\hline
 \multicolumn{2}{|c||}{Method} & \multicolumn{3}{c|}{Best-First} & \multicolumn{3}{c|}{Depth-First} & \multicolumn{3}{c|}{Breadth-First} & \multicolumn{3}{c|}{A$^*$} \\
\hline
 \multicolumn{2}{|c||}{$K$} & \textsc{$1$} & \textsc{$3$} & \textsc{$5$} & \textsc{$1$} &  \textsc{$3$} & \textsc{$5$} & \textsc{$1$} & \textsc{$3$} &  \textsc{$5$} & \textsc{$1$} & \textsc{$3$} &  \textsc{$5$} \\
\hline
\multirow{5}{*}{$N$} & $10$ & 100 & 100 & 100 & 100 & 100 & 100 & 100 & 35 & 0 & 100 & 100 & 35 \\
\cline{2-14}
& $15$ & 100 & 100 & 100 & 100 & 100 & 100 & 100 & 5 & 0 & 100 & 60 & 0 \\
\cline{2-14}
& $20$ & 100 & 100 & 100 & 100 & 100 & 100 & 100 & 0 & 0 & 100 & 35 & 0 \\
\cline{2-14}
& $25$ & 90 & 100 & 100 & 90 & 100 & 100 & 90 & 0 & 0 & 90 & 10 & 0 \\
\cline{2-14}
& $30$ & 40 & 85 & 100 & 50 & 90 & 100 & 45 & 0 & 0 & 55 & 0 & 0 \\
\hline
\end{tabular}}
\end{table}

We compare the repetitiveness of the four versions for $N = 10$. We do not consider $K=1$ since $\mathcal{O}_S$ is always filled with the same type of objects. Generally, Best-First and A$^*$ show lower values because they adopt the secondary cost to penalize consecutive objects belonging to the same group. 
With a larger $K$, it is natural to have more diversity in $\mathcal{O}_S$. From this result, we can expect that the execution time of Best-First and A$^*$ can be shorter than that of BFS and DFS because the depots are less congested, which will be confirmed in Sec.~\ref{sec:sim}.

\begin{figure}
   \centering
	\includegraphics[width=0.49\textwidth]{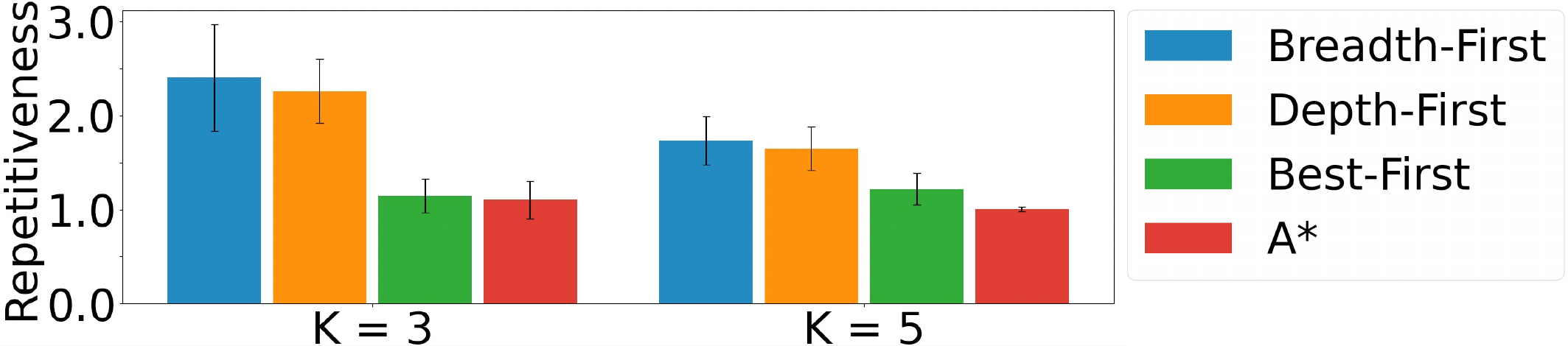}
	\caption{Comparison of the repetitiveness for $N=10$ and $K = 3, 5$}
    \label{fig:rep}
\vspace{-15pt}
\end{figure}

\subsection{Experiments in dynamic simulation}
\label{sec:sim}

We test Best-First and DFS which are shown to be fast enough for practical applications. Since extensive tests are already done in Sec.~\ref{sec:test}, we pick a subset of the test sets for this experiments in simulation. From the previously generated sets, we choose the sets with $N = 10, 20, 30$ and $K = 3, 5$ resulting in 120 instances in total. We modify the diameters of the disc objects such that the object with higher priority is the largest (30\,cm). This modification is to visually show the precedence constraint where a smaller object should be stacked on top of larger ones. We have three mobile manipulators ($M = 3$) as shown in Fig.~\ref{fig:unity}, which is in line with our assumption of $N > M$. The number of robots could vary but we fix it to limit the number of control variables, without loss of generality. 

Initially, $K$ depots are initialized. The depots and robots are located on the periphery of a clutter of objects.\footnote{The space limit precludes inserting large screenshots but the video clip at \url{https://youtu.be/oMBNWeFYGDw} should help.} Once Alg.~\ref{alg:sorting} finds $\mathcal{O}_S$, the robots start manipulating objects. The centralized task allocator assigns the object at the head of $\mathcal{O}_S$ to an idle robot. Each robot has its own global path planner and local controller for collision-free navigation and motion planner for manipulation. 
The robot picks and places an object using a suction gripper. Depending on the destination of the object (i.e., depot or buffer), which is described in an auxiliary list, the robot stacks it to its corresponding depot or temporarily unload it to a predefined buffer slot. If a robot is stacking an object to a depot, other robots that transport objects to the same depot must wait to keep the order constraint and avoid collisions. 

We measure metrics that only can be obtained if robots are used for physical (or simulated) sorting tasks. We measure task planning time for both computation of $\mathcal{O}_S$ and task allocation. 
Each sequence $\mathcal{O}_S$ is examined if the robots have feasible paths and motions to successfully sort all objects. We measure the time for motion planning to compute trajectories for navigation and manipulation. If no feasible motion or path is found at runtime, Alg.~\ref{alg:sorting} finds a new sequence with the remaining unsorted objects. This replanning time is also included in the task planning time. We also measure the execution time of robots from they start sorting til the end. All the results are shown in Table~\ref{tab:tamp}. 

\begin{figure}
    \centering
	\includegraphics[width=0.49\textwidth]{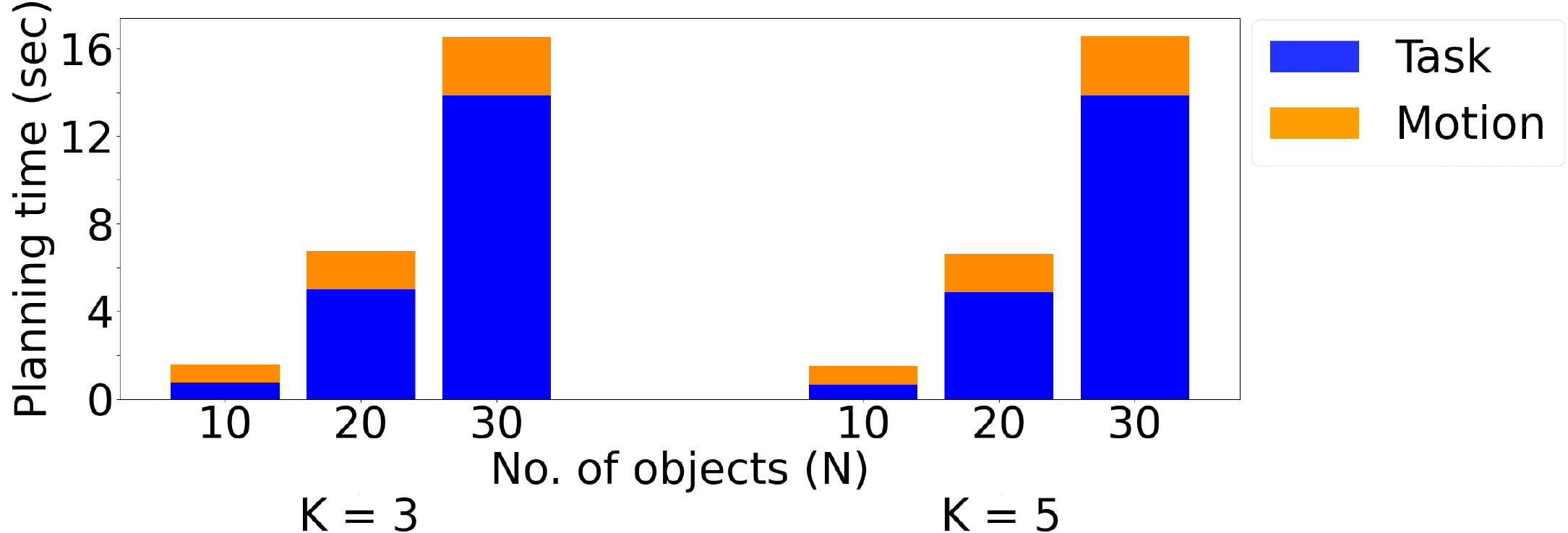}
	\caption{The total planning time of Best-First Search}
    \label{fig:sim_planning}
\end{figure}

\begin{table}[t!]
    \caption{The average task planning, motion planning, and execution time measured in the experiments in dynamic simulation}
            \centering
            \scalebox{0.83}{%
            \begin{tabular}{|c|c|c||c|c|c|c|}
            \hline
            \multicolumn{3}{|c||}{Measure (sec)} & \multicolumn{4}{c|}{Task planning time}\\
            \hline
             \multicolumn{3}{|c||}{Method} & \multicolumn{2}{c|}{Best-First} & \multicolumn{2}{c|}{Depth-First}\\
            \hline
             \multicolumn{3}{|c||}{$K$} & \textsc{$3$} & \textsc{$5$} & \textsc{$3$} & \textsc{$5$} \\
            \hline
             \multirow{6}{*}{$N$} & \multirow{2}{*}{$10$} & Mean & 0.7252 & 0.6468 & 0.7192 & 0.6495 \\
             \cline{3-7} && SD & 0.2032 & 0.0206 & 0.1944 & 0.0162 \\
             \cline{2-7} & \multirow{2}{*}{$20$} & Mean & 5.0579 & 4.9656 & 4.9774 & 4.8709 \\
             \cline{3-7} && SD & 0.0941 & 0.1009 & 0.1583 & 0.1300\\
             \cline{2-7} & \multirow{2}{*}{$30$} & Mean & 13.8530 & 14.9693 & 13.7532 & 13.8852 \\
             \cline{3-7} && SD & 0.2194 & 0.2670 & 0.2237 & 0.2219 \\
             \hhline{|=======|}
            \multicolumn{3}{|c||}{Measure (sec)} & \multicolumn{4}{c|}{Motion planning time}\\
            \hline
            \multicolumn{3}{|c||}{Method} & \multicolumn{2}{c|}{Best-First} & \multicolumn{2}{c|}{Depth-First} \\
            \hline
            \multicolumn{3}{|c||}{$K$} & \textsc{$3$} & \textsc{$5$} & \textsc{$3$} & \textsc{$5$} \\
            \hline
             \multirow{6}{*}{$N$} & \multirow{2}{*}{$10$} & Mean & 0.86689 & 0.8833 & 0.8493 & 0.8590 \\
             \cline{3-7} && SD &  0.0465 & 0.0386 & 0.0459 & 0.0496 \\
             \cline{2-7} & \multirow{2}{*}{$20$} & Mean & 1.7603 & 1.7635 & 1.7684 & 1.7527 \\
             \cline{3-7} && SD & 0.04760 & 0.0432 & 0.0446 & 0.0582\\
             \cline{2-7} & \multirow{2}{*}{$30$} & Mean & 2.7068 & 2.7076 & 2.6926 & 2.7033 \\
             \cline{3-7} && SD & 0.0545 & 0.0274 & 0.0456 & 0.0766\\
            \hhline{|=======|}
            \multicolumn{3}{|c||}{Measure (sec)} & \multicolumn{4}{c|}{Execution time}\\
            \hline
            \multicolumn{3}{|c||}{Method} & \multicolumn{2}{c|}{Best-First} & \multicolumn{2}{c|}{Depth-First} \\
            \hline
            \multicolumn{3}{|c||}{$K$} & \textsc{$3$} & \textsc{$5$} & \textsc{$3$} & \textsc{$5$} \\
            \hline
             \multirow{6}{*}{$N$} & \multirow{2}{*}{$10$} & Mean & 100.7069 & 101.2474 & 106.3651 & 101.0803 \\
             \cline{3-7} && SD &  4.6611 & 4.2495 & 6.7412 & 6.1202\\
             \cline{2-7} & \multirow{2}{*}{$20$} & Mean & 216.9901 & 213.0898 & 225.3280 & 220.1250 \\
             \cline{3-7} && SD & 9.3759 & 8.9444 & 8.1353 & 7.797\\
             \cline{2-7} & \multirow{2}{*}{$30$} & Mean & 331.7762 & 332.4469 & 351.8481 & 349.0364 \\
             \cline{3-7} && SD & 9.6318 & 12.2265 & 9.6746 & 9.4746\\
            \hline
             \end{tabular}}
     \label{tab:tamp}
     \vspace{-20pt}
\end{table}

Fig.~\ref{fig:sim_planning} shows the total planning time of Best-First consisting of task and motion planning time (Depth-First shows the same tendency). While the planning time of the two methods are similar, the execution time shows a meaningful difference. For example, $N = 30$ and $K = 3$ could incur more congestion around the depots. The difference of execution time in that pair is 19.4\,sec, which will increase if we operate more robots or have a smaller $K$. It shows that the secondary cost presented in Sec.~\ref{sec:cost} is effective.

\section{Conclusion}

We consider the problem of coordinating multiple mobile manipulators to sort objects in a particular order while the order cannot be satisfied unless some objects are rearranged to access the objects to be sorted. To solve this computationally intractable problem, we present a search-based method with four different search strategies. They are shown to be complete where two of them are shown to produce solutions fast enough. We demonstrate our methods handle both monotone and nonmonotone rearrangement instances while securing high success rate by interactive seeking for feasible tasks through task and motion planning method. With this work as a baseline, we would develop further by generalizing object geometry. Also, we seek to extend our work to be used in other applications.

\bibliographystyle{IEEEtran}
\bibliography{references}

\end{document}